\documentclass[doubleblind]{ceurart}

\usepackage{amsmath, amssymb, amsthm, amsfonts}
\usepackage{mathtools}
\usepackage{bm}

\usepackage{multirow}
\usepackage{array}
\usepackage{enumitem}
\setlist{nosep,leftmargin=1.2em}

\usepackage{ragged2e}

\usepackage{listings}
\usepackage{xcolor}
\usepackage{tikz}
\usetikzlibrary{positioning, arrows.meta, shapes.geometric, fit, backgrounds, calc, decorations.pathreplacing}

\definecolor{nodeblue}{HTML}{D7E4F5}
\definecolor{nodegreen}{HTML}{D9ECD0}
\definecolor{nodeorange}{HTML}{FCE5CD}
\definecolor{nodepurple}{HTML}{E1D5F0}
\definecolor{nodegrey}{HTML}{ECECEC}
\definecolor{edge}{HTML}{3A3A3A}

\tikzset{
  stage/.style    = {rectangle, rounded corners=2pt, draw=edge, line width=0.4pt,
                     fill=nodeblue, align=center, inner sep=5pt, font=\small},
  retr/.style     = {stage, fill=nodegreen},
  llm/.style      = {stage, fill=nodeorange},
  decide/.style   = {diamond, aspect=2, draw=edge, line width=0.4pt,
                     fill=nodepurple, align=center, inner sep=2pt, font=\small},
  io/.style       = {stage, fill=nodegrey},
  flow/.style     = {-{Latex[length=2.2mm]}, draw=edge, line width=0.5pt},
  flowdash/.style = {flow, dashed},
  grouplbl/.style = {font=\footnotesize\itshape, text=edge!80}
}

\theoremstyle{plain}
\newtheorem{theorem}{Theorem}
\newtheorem{lemma}[theorem]{Lemma}

\theoremstyle{definition}
\newtheorem{definition}[theorem]{Definition}
\theoremstyle{remark}
\newtheorem{observation}[theorem]{Observation}

\newcommand{\score}{\operatorname{score}}
\newcommand{\heads}{\mathcal{H}}

\newcommand{\alwayspick}[1]{\textsc{always-}#1}
\newcommand{\E}{\mathbb{E}}

\begin{document}

\copyrightyear{2026}
\copyrightclause{Copyright for this paper by its authors.
  Use permitted under Creative Commons License Attribution 4.0
  International (CC BY 4.0).}

\conference{CLEF 2026 Working Notes, 21--24 September 2026, Jena, Germany}

\title{Cost-Pragmatic Quality Gating and Selection--Fusion Multi-Model
  Combiners for BioASQ Phase A+/B}

\title[mode=sub]{BioASQ Task 14B 2026}

\author[1]{Dima Galat}[%
orcid=0000-0003-3825-2142,
email=dima.galat@student.uts.edu.au,
]
\cormark[1]

\author[1]{Marian-Andrei Rizoiu}[%
orcid=0000-0003-0381-669X,
email=Marian-Andrei.Rizoiu@uts.edu.au,
]

\address[1]{University of Technology Sydney, Australia}

\cortext[1]{Corresponding author.}

\begin{abstract}
We describe our BioASQ Task~14B 2026 system. The work centres on two
design decisions: how aggressively to re-retrieve when first-stage
retrieval is weak, and how to combine multiple language-model answers.
Retrieval unions two parallel pipelines --- a hybrid first stage
(dense BGE + BM25 + RRF, reaching R@200 = 99.3\% on the
BioASQ-13b historical archive) and an agent-driven pipeline that
decomposes the question over PubMed, Europe~PMC, and iCite --- with a
BGE cross-encoder quality gate flagging weakly-supported questions for
selective re-retrieval. On Task~12B 2024 validation, a cost-pragmatic
re-retrieval policy beats a skill-strict baseline significantly on
list~$F_1$ and list precision, at $\approx 12\%$ lower re-retrieval
cost. Holding prompt and model fixed across val and
test~13B (different question sets), list~$F_1$ rises by $+0.132$
absolute on the BioASQ-released gold-input pool, consistent with
substantial retrieval-side headroom. For Phase~B
answering we decompose multi-model
ensemble lift into a \emph{selection} component bounded by the
per-question oracle and a \emph{fusion} component that aggregators can
exceed. The decomposition predicts \emph{before} any experiment that
LLM-as-judge wins on selection-dominated metrics (yes/no, multi-reference
ROUGE) but is structurally insufficient on the recall component of
fusion-friendly metrics (factoid rank-1, list recall). On Task~13B 2025
our synonym-union resolver wins list recall on every head, while GPT-5.5
solo retains the list-$F_1$ lead because the resolver's wider item set
costs precision. On the Task~14B 2026 preliminary leaderboard our team
places \emph{first} on the combined-exact aggregate on three of the
eight (phase~$\times$~batch) leaderboards, wins four individual
question-type cells, and takes \#1 on Phase~B b3 ideal.
\end{abstract}


\maketitle

\section{Introduction}
\label{sec:intro}
The BioASQ task-B challenge has run since 2013 \citep{tsatsaronis2015bioasq}
and now covers four answer types (yes/no, factoid, list, and summary)
over a corpus drawn from PubMed \citep{krithara2023bioasqqa}. Modern
systems have largely converged on a retrieval-augmented multi-stage
architecture \citep{lewis2020rag, izacard2021fid}: candidate retrieval
over PubMed via E-utilities and dense/sparse indices, reranking with
cross-encoders, and prompt-engineered language-model (LM) answering
over the retrieved context. The competitive surface is no longer ``do we have a
language model'' but \emph{how much retrieval is enough, and which combiner
is right for which metric}.

Our work makes four contributions to that surface.

\begin{enumerate}[topsep=2pt, itemsep=3pt]
\item \textbf{A quantitative cost--quality decomposition of agent
re-retrieval.} Iterative agent retrieval (concept decomposition $\to$
multi-strategy queries $\to$ citation expansion) is expensive at the
token-budget level. We quality-gate every question with a BGE cross-encoder,
partition flagged questions into Tier~1 (severe coverage failure) / Tier~2 /
Tier~3 (borderline), and run a controlled ablation of three re-retrieval
policies on the full Task~12B 2024 validation set ($n=340$). The
cost-pragmatic policy that re-retrieves only Tier~1 plus selectively
rescues Tier~2 saves $\approx 12\%$ re-retrieval cost while significantly
beating the skill-strict baseline on list~$F_1$ (paired bootstrap CI
$[+0.004, +0.049]$) and list precision, and remaining non-inferior on
list recall.

\item \textbf{A selection--fusion decomposition of multi-model ensemble lift,
with metric-structure-driven combiner choice.} Multi-model ensembling can
combine candidates either by \emph{selection} (an LLM-as-judge returns one
verbatim) or by \emph{aggregation} (a deterministic rule constructs a new
candidate, e.g., by unioning synonym sets). We show
(Lemma~\ref{lem:judge-ceiling}) that any selector is bounded above by the
per-question oracle, while a synonym-union aggregator can exceed the oracle
on metrics that reward candidate-set enlargement. The decomposition
predicts \emph{before} any experiment that LLM judging is sufficient on
selection-dominated metrics
(yes/no, multi-reference ROUGE) but structurally insufficient on the recall
component of fusion-friendly metrics (factoid rank-1 against gold synonyms; list
recall with synonym matching). Whether this recall lift translates to
list~$F_1$ is precision-conditional: \S\ref{sec:exp:test13} reports that the
resolver wins list recall but cedes list~$F_1$ to GPT-5.5 solo on Task~13B.
An \emph{always-pick floor} (Observation~\ref{obs:always-pick}) gives
a trivially-computable diagnostic that any deployed judge must clear.

\item \textbf{A val\,$\to$\,test-gold gap that separates
retrieval-bound from answer-bound metrics.} Held-out Task~13B 2025
numbers on the BioASQ-released gold-input pool put yes/no at the answer
ceiling already ($\Delta$ vs val $= +0.004$) while list~$F_1$ jumps by
$+0.132$ absolute. We hold prompt and model fixed across val and
test~13B (\S\ref{sec:exp:test13}), but the question sets differ, so
the gap is best read as \emph{consistent with substantial retrieval-side
headroom} rather than as a clean retrieval-only attribution. Closing
the gap would require running our retrieval on the test~13B question
set.

\item \textbf{A synonym-union resolver that exploits BioASQ's synonym-aware
list scoring.} Following the framework of~(2), our list combiner is not an
LLM judge but a deterministic synonym-union aggregator across Claude,
Gemini, and GPT-5.5 list answers. It gains significant list recall on
Task~13B 2025 ($+0.025$, paired bootstrap CI~$[+0.009, +0.045]$) and
wins the Phase~B b3 ideal track \#1 on the Task~14B 2026 leaderboard.

\end{enumerate}

\noindent\textbf{Headline result.} On the Task~14B 2026 preliminary
leaderboard our system places \emph{first} on the combined-exact
aggregate (mean of Y/N Macro~$F_1$, factoid MRR, and list~$F_1$) on
three of the eight (phase~$\times$~batch) leaderboards, with
second-place placements on three further leaderboards.

\noindent
The remainder of the paper is organised as follows. \S\ref{sec:related}
reviews related work; \S\ref{sec:retrieval}--\ref{sec:reretrieval} give a
system overview and describe the two-pipeline retrieval architecture, the
BGE quality gate, and the cost-quality
re-retrieval ablation; \S\ref{sec:answering} introduces the four Phase~B
combiners including the selection--fusion framework; \S\ref{sec:experiments}
presents results on validation, held-out test, and the live Task~14B
leaderboard; \S\ref{sec:discussion} discusses implications and limitations;
\S\ref{sec:conclusion} concludes.

\section{Related work}
\label{sec:related}

\paragraph{BioASQ Task B.}
The BioASQ challenge \citep{tsatsaronis2015bioasq, krithara2023bioasqqa,
nentidis2024bioasq12, nentidis2025bioasq13, nentidis2026bioasq14} has run since 2013 as the
canonical biomedical-QA benchmark. Adjacent biomedical-QA evaluations
include PubMedQA \citep{jin2019pubmedqa} (snippet-level closed-domain
QA) and MedQA / USMLE \citep{jin2021medqa} (medical-exam multiple
choice), but BioASQ remains unique in evaluating four answer types
end-to-end over PubMed retrieval. Task~B asks systems to retrieve relevant PubMed abstracts and extract
supporting snippets (Phase~A), and produce four answer types (yes/no,
factoid, list, summary) from those snippets (Phase~B). The official
evaluation uses an exact-answer metric per type (accuracy / MRR / $F_1$)
plus ROUGE-2/SU4 \citep{lin2004rouge} on free-text ``ideal'' answers; manual
ideal-answer scores are released after the workshop. Participation has
shifted from classical IR + template answering through BERT-style
biomedical encoders \citep{lee2020biobert} to instruction-tuned LM
answering over retrieval-augmented context. We evaluate on Task~12B 2024
(validation), Task~13B 2025 (held-out test), and submit live on
Task~14B 2026 \citep{nentidis2026task14b} (\S\ref{sec:exp:live}).

\paragraph{Multi-stage and agentic retrieval.}
Modern biomedical retrieval combines a high-recall first stage
(BM25~\citep{robertson2009okapi}, dense bi-encoders such as
DPR~\citep{karpukhin2020dpr}, Contriever~\citep{izacard2022contriever},
ColBERT~\citep{khattab2020colbert}, or domain-specific
MedCPT~\citep{jin2023medcpt}) with a high-precision cross-encoder
reranker~\citep{nogueira2020monot5,chen2024bgem3}, often fused with
reciprocal-rank fusion \citep{cormack2009rrf} and evaluated on
heterogeneous benchmarks such as BEIR~\citep{thakur2021beir}. We use the BGE family \citep{chen2024bgem3, xiao2024cpack}
(specifically \texttt{bge-base-en-v1.5} from C-Pack and
\texttt{bge-reranker-v2-m3} from the M3 series) as both the dense
first-stage encoder and the quality-gate cross-encoder, and report a
controlled study showing that cross-encoder reranking applied on top
of an already-fused biomedical first stage \emph{degrades} R@10
(Appendix~\ref{app:retrieval-studies}). Iterative LM-driven retrieval ---
question decomposition, multi-query, citation expansion, triage, builds
on ReAct \citep{yao2023react} and Toolformer
\citep{schick2023toolformer}; PubMed-specific synonym/MeSH expansion
follows MetaMap \citep{aronson2010metamap}, and citation expansion uses
iCite~\citep{hutchins2019icite}.

\paragraph{Biomedical LM answering and multi-model ensembling.}
Modern biomedical answering uses instruction-tuned LMs
\citep{ouyang2022instructgpt} over retrieved context, building on the
Transformer architecture~\citep{vaswani2017attention} and the
BERT-style masked-LM~\citep{devlin2019bert} and few-shot GPT-style
prompted-LM~\citep{brown2020gpt3} paradigms. Domain-specific encoders
(SciBERT~\citep{beltagy2019scibert},
PubMedBERT~\citep{gu2021pubmedbert}) and domain-tuned generative models
(BioGPT~\citep{luo2022biogpt}, Med-PaLM~\citep{singhal2023medpalm})
gave clear early gains over general models but are overtaken by
sufficiently large general models with good retrieval, so we use
general-purpose models without domain-specific fine-tuning. Our
single-agent prompt design (\S\ref{sec:answering:prompt}) builds on chain-of-thought
prompting \citep{wei2022cot} by structuring reasoning per answer type
rather than as free-form rationales.
The classical theory of combining classifiers
\citep{kittler1998combining} already distinguishes \emph{selectors}
(max, voting, single-pick judges) from \emph{combiners} (sum, product,
weighted average); our framework adapts that taxonomy to natural-language
outputs by treating synonym-union as a combiner. Multi-model ensembling
of large language models has three contemporary paradigms: voting and
self-consistency~\citep{wang2023selfconsistency}; probability-weighted
aggregation (requires calibrated log-probabilities across heads);
LLM-as-judge~\citep{zheng2023llmjudge}, which uses a learned model to
select one candidate. Recent extensions include pairwise-ranked
ensembling with generative fusion across heads
\citep{jiang2023llmblender}, multi-agent debate
\citep{du2023llmdebate}, and cross-examination
\citep{cohen2023lmvslm} that orchestrate multi-turn LM interactions to
improve factuality. The first three (voting, weighted aggregation,
judging) are selector-style in our taxonomy and therefore subject to
the judge ceiling (Lemma~\ref{lem:judge-ceiling}); LLM-Blender's
generative-fusion module is the closest published analogue to our
synonym-union resolver, with the difference that it is trained
end-to-end rather than constructed deterministically from synonym sets. Our system intentionally combines list answers with
a deterministic synonym-union aggregator rather than an LLM judge: any
LLM-as-judge returning one candidate verbatim is a selector, and its
expected score is bounded above by the per-question selection oracle.
When the target metric rewards candidate-set enlargement, aggregators
can produce outputs no selector can match. \S\ref{sec:answering:syn}
formalises this.

\section{System overview and retrieval pipeline}
\label{sec:overview}
\label{sec:retrieval}

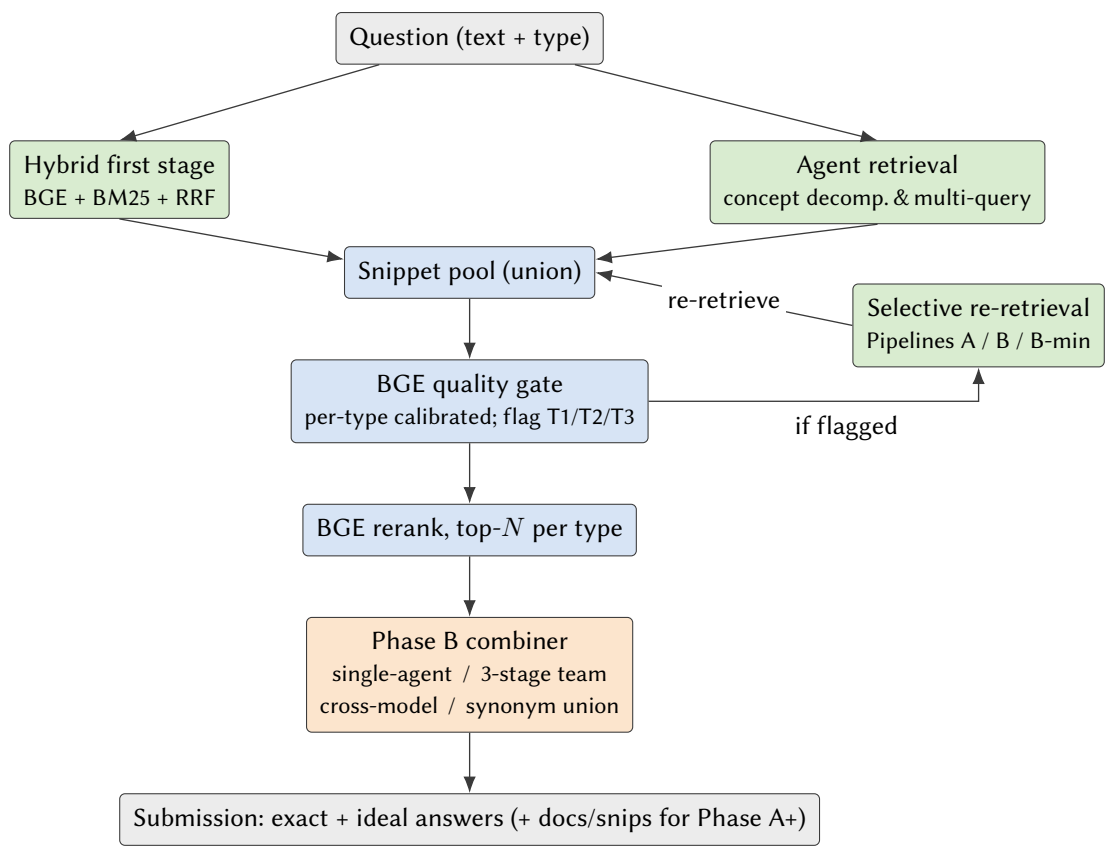
\begin{figure*}[t]
\centering
\begin{tikzpicture}[node distance=8mm and 12mm]
  \node[io] (qin) {Question (text + type)};
  \node[retr, below left=10mm and 14mm of qin] (hybrid)
    {Hybrid first stage\\\footnotesize BGE + BM25 + RRF};
  \node[retr, below right=10mm and 14mm of qin] (agent)
    {Agent retrieval\\\footnotesize concept decomp.\,\&\,multi-query};
  \node[stage, below=24mm of qin] (union) {Snippet pool (union)};
  \node[retr, right=34mm of union, yshift=-7mm] (reretr)
    {Selective re-retrieval\\\footnotesize Pipelines A / B / B-min};
  \node[stage, below=of union] (gate)
    {BGE quality gate\\\footnotesize per-type calibrated; flag T1/T2/T3};
  \node[stage, below=of gate] (rerank) {BGE rerank, top-$N$ per type};
  \node[llm, below=of rerank] (answer)
    {Phase~B combiner\\\footnotesize single-agent \,/\, 3-stage team\\
     \footnotesize cross-model \,/\, synonym union};
  \node[io, below=of answer] (out)
    {Submission: exact + ideal answers (+ docs/snips for Phase~A+)};

  \draw[flow] (qin) -- (hybrid.north);
  \draw[flow] (qin) -- (agent.north);
  \draw[flow] (hybrid.south) -- ($(union.north west)!0.5!(union.west)$);
  \draw[flow] (agent.south)  -- ($(union.north east)!0.5!(union.east)$);
  \draw[flow] (union) -- (gate);
  \draw[flow] (gate)  -- (rerank);
  \draw[flow] (rerank) -- (answer);
  \draw[flow] (answer) -- (out);
  \draw[flow] (gate.east) -| node[pos=0.3, below=3pt, fill=white,
              inner sep=2pt, font=\small] {if flagged} (reretr.south);
  \draw[flow] (reretr.west) -- node[midway, fill=white, inner sep=2pt,
              font=\small] {re-retrieve} (union.east);
\end{tikzpicture}
\caption{\justifying System pipeline. Two parallel retrievers (green) feed a unioned
snippet pool that the BGE quality gate scores. Flagged questions trigger
selective re-retrieval whose output rejoins the union; the rest flows
down through BGE rerank, the Phase~B combiner (orange), and submission.
The hybrid first stage and the BGE gate share the same encoder family
but operate at different layers: the first stage runs over the entire
corpus, the gate over per-question candidates.}
\label{fig:overview}
\end{figure*}

Figure~\ref{fig:overview} shows the system. Two retrieval pipelines run
in parallel: a hybrid first stage (BGE~+~BM25~+~RRF;
\S\ref{sec:retrieval:hybrid}) and an LM-driven agent pipeline
(concept decomposition + multi-query + iCite expansion;
\S\ref{sec:retrieval:agent}), with their outputs unioned. A BGE
cross-encoder quality gate (\S\ref{sec:gate}) scores every snippet and
flags questions for selective re-retrieval (\S\ref{sec:reretrieval}).
The post-gate snippet pool feeds four Phase~B combiners
(\S\ref{sec:answering}) which produce the exact and ideal answers
submitted to BioASQ.

Phase~B answering uses Claude Opus~4.6 \citep{anthropic2026claude},
Gemini~2.5~Pro \citep{google2026gemini}, and GPT-5.5
\citep{openai2026gpt55}; the BGE reranker is served locally via
\texttt{llama.cpp}.

\subsection{Hybrid first stage}
\label{sec:retrieval:hybrid}

Two complementary first-stage retrievers run in parallel and are fused with
reciprocal-rank fusion (RRF) \citep{cormack2009rrf} with $k = 10$, returning
the top-200. We deliberately omit a cross-encoder reranker downstream of the
first stage; Appendix~\ref{app:retrieval-studies} reports the controlled study
that motivates this choice.

\paragraph{Corpus and indices.}
The retrieval corpus is a curated PubMed slice of $\approx 58{,}000$
documents (BioASQ-13b years~5--13). Each document carries the
\emph{full abstract} from NCBI E-utilities; using full abstracts rather
than BioASQ snippet-only text is the single most consequential design
decision, lifting recall from a snippet-only ceiling of $\approx 82\%$
R@200 to $\approx 96\%$. The pipeline materialises parallel dense
(HNSW~\citep{malkov2020hnsw}) and BM25 indices on disk; the production
dense encoder is BAAI/\texttt{bge-base-en-v1.5}~\citep{chen2024bgem3}.

On the historical BioASQ-13b retrieval slice (3{,}986 questions, years
5--13) the first stage reaches R@200 = 99.3\% with sub-100~ms
query latency. Appendix~\ref{app:retrieval-studies} reports the full
benchmark, the per-component ablation showing that hybrid fusion is
what lifts recall past the $\approx 80$--$83\%$ BM25-only and
semantic-only plateaus, and the intuition for why complementary
first-stage signals matter on biomedical queries.

\paragraph{Encoder and reranker choices justified empirically.}
The production encoder was selected from eleven candidates benchmarked
under an identical pipeline. \texttt{BGE-large} (1024-dimensional)
\emph{underperformed} \texttt{BGE-base} on this corpus; domain-specific
biomedical encoders (MedCPT, MedEmbed) come close but do not surpass
general contrastive training because the corpus enrichment above
already closes the vocabulary gap. We further deliberately omit a cross-encoder
reranker downstream of the first stage: all four cross-encoders
evaluated \emph{degraded} R@10 (by 3--24\%), plausibly because their
web-style training distribution demotes relevant biomedical abstracts
that the RRF stage has already correctly surfaced
(Table~\ref{tab:rerankers}). The full encoder benchmark is tabulated
in Appendix~\ref{app:retrieval-studies} (Table~\ref{tab:embeddings}).
The same BGE cross-encoder we omit from the first stage is used
downstream as the per-question quality gate (\S\ref{sec:gate}), where
its role is calibrated filtering rather than corpus-wide reranking.

\begin{table*}[t]
\centering
\small
\caption{Cross-encoder reranker results, applied on top of the hybrid
first stage. $\Delta$R@10 is the \emph{relative} change (\% off the
no-rerank baseline of $0.687$). Negative is worse.}
\label{tab:rerankers}
\begin{tabular}{lrl}
\toprule
Reranker & $\Delta$R@10\,$\uparrow$ & Verdict \\
\midrule
BAAI/bge-reranker-base                  & $-24.4\%$ & Catastrophic \\
BAAI/bge-reranker-large                 & $-19.6\%$ & Catastrophic \\
ncbi/MedCPT-Cross-Encoder               & $-3.7\%$  & Hurts \\
cross-encoder/ms-marco-MiniLM-L-6-v2    & $-3.1\%$  & Hurts \\
\bottomrule
\end{tabular}
\end{table*}

\subsection{Agent retrieval}
\label{sec:retrieval:agent}

The agent reasons explicitly about a question's key terms, synonyms,
and broader concepts before issuing queries. The most consequential step is
\emph{concept decomposition}: enumerating MeSH headings, synonyms,
abbreviations, broader/narrower terms, and British/American spelling
variants before any query is issued. Skipping it (even with strong
downstream search) costs $\approx 0.10$ list-$F_1$ on validation. The
agent then issues a portfolio of PubMed queries (MeSH-tagged,
free-text synonym sweeps, broader/recency), Europe~PMC full-text
queries \citep{levchenko2018europepmc} for terms unlikely to appear in
title/abstract, and iCite backward + forward citation expansion
\citep{hutchins2019icite} from high-relevance seeds. Each candidate is
triaged and at most one snippet sentence is extracted per relevant
article, leaving 15--20 snippets per question in the final output.

\subsection{Ensemble of hybrid and agent retrieval}
\label{sec:retrieval:ensemble}

For Task~14B 2026 batches~b3 and~b4 we run both retrievers and union their
snippet outputs. Snippet overlap on the same question is $\approx 4.5\%$,
i.e.\ the two sources are nearly disjoint. The union enters the BGE
quality gate (\S\ref{sec:gate}); the gate's rerank-and-top-$N$ cap
prevents the union from inflating the per-question pool size. On~b1 and~b2
we ran the agent alone (the hybrid first stage was integrated into the
Phase~A+ pipeline only mid-challenge), which gives a within-team natural
comparison: agent-only on the first two batches, agent~$\cup$~hybrid on the
last two.

\section{BGE quality gate}
\label{sec:gate}

Every Phase~A output is scored sentence-by-sentence with the BGE
cross-encoder. We aggregate per question into two signals (the
top-scoring snippet and the number of snippets above a per-type
threshold calibrated on validation) and flag a question when either
indicates retrieval failure. Flagged questions are tiered as \textbf{T1}
(severe coverage failure), \textbf{T2} (weak coverage), and \textbf{T3}
(borderline). Flag rates ranged $23$--$25\%$ across val and b1--b3,
with b4 a $40\%$ outlier: the densest batch on entity count and
disproportionately weak on retrieval against then-recent PubMed
indexing (\S\ref{sec:exp:live}).

\section{Re-retrieval and the cost--quality ablation}
\label{sec:reretrieval}

For each flagged question we generate a \emph{concept-decomposition
helper}: a one-shot LM analysis that explicitly names the retrieval gap
and writes mandatory queries. A re-retrieval sub-agent then reads the
helper output and runs a fresh search seeded by those queries.
Re-retrieval is more expensive per-question than the original
retrieval, so the policy question is \emph{which flagged
questions to re-retrieve}. We compare three policies, all sharing the
Phase~B prompt (\S\ref{sec:answering:prompt}):
\begin{itemize}[topsep=2pt]
\item \textbf{Pipeline~A (skill-strict)}: re-retrieve every Tier~1 and
Tier~2 flagged question (42 on validation).
\item \textbf{Pipeline~B-min (cost-pragmatic, no rescue)}: re-retrieve
only Tier~1 (15 on validation); Tier~2 and Tier~3 keep BGE-reranked
original snippets.
\item \textbf{Pipeline~B (cost-pragmatic with rescue)}: B-min plus a
\emph{rescue step} that re-retrieves Tier~2 questions whose
post-rerank top-1 remains weak; this brings the total to 37
re-retrievals on validation.
\end{itemize}
Re-retrieval results union with original-pool snippets, the BGE gate
re-scores, and we keep top-$N$ per type (15 factoid; 20 yes/no, list,
summary).

\subsection{Full ablation ($n=340$)}
\label{sec:reretrieval:full}

Table~\ref{tab:ablation} reports the full ablation. The non-inferiority
margin is $-0.005$ on the lower bound of the paired bootstrap CI
\citep{koehn2004bootstrap} ($n_\text{iter}=2000$). Per-type counts: yes/no~$=102$, factoid~$=85$,
list~$=80$, summary~$=73$. ROUGE-2 and ROUGE-SU4 are computed by our
standalone evaluator on bigram and skip-bigram (max-skip~4) $F$-measure,
taking max over multiple gold references where present
\citep{lin2004rouge}; absolute numbers differ from the official Perl
ROUGE-1.5.5 scorer because of the max-vs-average convention, but
within-system $\Delta$s are valid.

\begin{table*}[t]
\centering
\small
\caption{Re-retrieval policy ablation on Task~12B 2024 validation ($n=340$).
A = skill-strict (re-retrieve all 42 flagged); B = cost-pragmatic with
rescue (37); B-min = cost-pragmatic without rescue (15). Bold = winner per
row. NI~$=$~non-inferiority verdict at margin $-0.005$ with paired
bootstrap 95\% CI ($n_\text{iter}=2000$).}
\label{tab:ablation}
\begin{tabular}{lcccll}
\toprule
Metric & A & B & B-min & NI: B vs A & NI: B-min vs A \\
\midrule
YN accuracy             & 0.9216 & 0.9216 & 0.9216 & tie (0/102) & tie (0/102) \\
Factoid MRR             & 0.6165 & 0.6106 & \textbf{0.6386} & tie ($\Delta -0.006$) & $+0.022$ \,$[-0.017, +0.063]$ \\
Factoid strict          & 0.5765 & 0.5765 & \textbf{0.6118} & tie & $+0.035$ \,$[-0.024, +0.094]$ \\
Factoid lenient         & 0.6941 & 0.6824 & 0.6941 & tie & tie \\
\textbf{List $F_1$}     & 0.4477 & \textbf{0.4733} & 0.4579 & \textbf{PASS} $[+0.004, +0.049]$ & tie $[-0.014, +0.036]$ \\
\textbf{List recall}    & 0.5051 & \textbf{0.5192} & 0.5106 & \textbf{PASS} $[-0.004, +0.034]$ & tie \\
\textbf{List precision} & 0.4438 & \textbf{0.4737} & 0.4596 & \textbf{PASS} $[+0.004, +0.058]$ & tie \\
ROUGE-2 F               & \textbf{0.2601} & 0.2543 & 0.2547 & $\Delta -0.006$ & $\Delta -0.005$ \\
ROUGE-SU4 F             & \textbf{0.2498} & 0.2447 & 0.2466 & $\Delta -0.005$ & $\Delta -0.003$ \\
\bottomrule
\end{tabular}
\end{table*}

\paragraph{Findings.}
Pipeline~B is significantly better than~A on list~$F_1$ and list
precision (CIs strictly positive); on list recall the CI
$[-0.004,+0.034]$ passes the $-0.005$ non-inferiority margin but
crosses zero, so the result is non-inferior rather than superior. The
rescue step accounts for the gain, and without it list~$F_1$ reverts
to within-margin. B-min leads factoid point estimates but at $n=85$
the CIs cross zero, so the non-inferiority bound fails (a replicator
should plan for $n\!\ge\!150$). YN is fully retrieval-robust: all
three pipelines produced identical answers on every validation
question. Pipeline~B saves $\approx 12\%$ of re-retrieval cost over~A
while winning lists significantly; Pipeline~B-min saves $\approx 64\%$
with no significant difference from~A on any metric, but its CIs at
this~$n$ are too wide to establish non-inferiority within the
$-0.005$ margin on factoid~MRR or list~$F_1$. Pipeline~B is our
production setting.

\section{Phase B answering}
\label{sec:answering}

We compare four Phase~B combiners. \S\ref{sec:answering:prompt} introduces the
Claude solo prompt that anchors all variants;
\S\ref{sec:answering:team} describes a 3-stage QA team;
\S\ref{sec:answering:xm} describes cross-model routing; and
\S\ref{sec:answering:syn} introduces the synonym-union resolver, together
with its theoretical grounding in a selection--fusion decomposition.

\subsection{The Claude solo prompt}
\label{sec:answering:prompt}

Our baseline prompt is the result of 10+ A/B-test iterations on
validation and comprises four design choices: a \textbf{four-category
yes/no framework} (A association, B clinical benefit, C approval /
safety / standard, D factual claim) that prefixes the
\texttt{ideal\_answer} with the category letter; \textbf{mandatory five
factoid candidates with 4--6 synonym variants each}; \textbf{two-stage
list extraction with per-item synonym expansion and a coverage re-read};
and \textbf{verbatim summary phrasing at 35--50 words} to maximise
ROUGE-2 bigram overlap with gold. Detailed descriptions are in
Appendix~\ref{app:claude-prompt}.

\subsection{Three-stage QA team}
\label{sec:answering:team}

The 3-stage QA team is a more structured alternative to the single-agent
prompt, with separate \emph{Evidence Analyst}, \emph{Answer Specialist},
and \emph{Validator} agents in the spirit of self-refinement
\citep{madaan2023selfrefine} and chain-of-verification
\citep{dhuliawala2023cove} but with explicit role separation rather
than iterative self-feedback. The Analyst produces a structured
evidence brief; the Specialist applies the single-agent rules using the brief;
the Validator cross-checks format compliance and content against
snippets. The team gains list recall on noisy retrieval but loses
factoid MRR (the Analyst's compression strips the gestalt judgement of
which entity \emph{is} the answer vs.\ which is merely discussed); the
test-13B numbers in \S\ref{sec:exp:test13} make this trade-off
explicit.

\subsection{Cross-model routing}
\label{sec:answering:xm}

Cross-model routing assigns each question type to a head from a
\emph{different model family}: factoid, yes/no, and summary go to
Claude solo; list questions go to a Gemini~2.5~Pro 3-stage team. The motivation is per-metric
specialisation, in the spirit of pairwise-ranked LLM ensembling
\citep{jiang2023llmblender} but with explicit per-question-type
routing rather than a learned ranker over candidates. Gemini's
structured-output bias plays well with the explicit
Analyst~$\to$~Specialist chain, and the routed configuration is the
best validated overall on lists; the full test-13B comparison appears
in \S\ref{sec:exp:test13}.

\subsection{Synonym-union resolver}
\label{sec:answering:syn}

For list answers we use a deterministic synonym-union aggregator rather
than an LLM judge. We first give the intuition with a worked example,
then the decomposition that motivates the design, the BioASQ track
taxonomy it induces, a cheap deployment diagnostic, and the
implementation.

\paragraph{A worked example.}
\label{sec:answering:syn:decomp}
Consider a list question whose gold answer is $\{A, C\}$ and two heads
that answer $\{A, B\}$ and $\{B, C\}$. Each head alone recovers exactly
one gold item (recall $1/2$), and an LLM-as-judge --- which must return
one head's answer verbatim --- is therefore capped at recall $1/2$
whichever head it picks. The union $\{A, B, C\}$ recovers \emph{both}
gold items (recall $2/2$) but pays for it in precision by admitting the
spurious~$B$. That is the whole message: on a metric that rewards a
larger correct candidate set, a rule that \emph{unions} candidates can
beat any rule that \emph{selects} one of them. The resolver of
\S\ref{sec:answering:syn:impl} is exactly this union step, which is also
why it wins list recall but cedes list~$F_1$ --- the precision cost ---
in \S\ref{sec:exp:test13}. The formalism below only makes ``rewards a
larger candidate set'' precise enough to predict, per track, whether the
judge or the union is the right combiner.

\paragraph{Selection lift vs.\ fusion lift.}

Fix a finite set of questions~$Q$, a finite set of answer-generation
heads~$\heads$, and a per-question metric
$\score : \mathcal{A} \times \mathcal{A} \to \mathbb{R}_{\ge 0}$ on an
answer space~$\mathcal{A}$. Let $a_{h,q}$ denote head~$h$'s answer to
question~$q$ and $g(q)$ the gold reference. Write
$S(h) = \E_q[\score(g(q), a_{h,q})]$ for head~$h$'s population score;
all expectations~$\E_q$ are empirical averages over the evaluation
set~$Q$ unless otherwise stated.

Two combiner families are available, paralleling the classical
selector / fuser dichotomy in classifier-combination theory
\citep{kittler1998combining}. A \emph{selector}~$J$ returns one of
the candidates verbatim: $J(q) \in \heads$. An \emph{aggregator}~$\Phi$
produces an answer $a_{\Phi,q}$ not necessarily equal to any~$a_{h,q}$ ---
for example, by unioning their synonym sets.

\begin{definition}[Selection oracle]
The selection oracle~$\Omega^*$ picks a per-question best head (ties
broken arbitrarily):
\begin{align*}
\Omega^*(q) &\,\in\, \arg\max_{h \in \heads}\, \score(g(q), a_{h,q}), \\
S(\Omega^*) &\,=\, \E_q\!\left[\,\max_{h \in \heads} \score(g(q), a_{h,q})\,\right].
\end{align*}
$S(\Omega^*)$ does not depend on the tie-breaking rule. The
\emph{selection lift} of~$\heads$ is
$\Delta_{\mathrm{sel}}(\heads) = S(\Omega^*) - \max_h S(h) \ge 0$.
\end{definition}

\begin{definition}[Aggregator]
An aggregator is a function~$\Phi$ mapping the candidate tuple
$(a_{h,q})_{h \in \heads}$ to an answer $a_{\Phi,q} \in \mathcal{A}$ not
necessarily equal to any~$a_{h,q}$. The \emph{fusion lift} of~$\Phi$ is
$\Delta_{\mathrm{fus}}(\Phi, \heads) = S(\Phi) - S(\Omega^*)$, which can be
either sign.
\end{definition}

Selection lift is non-negative; fusion lift can be positive (a
synonym-union aggregator that enlarges a rank-1 set will typically have
$\Delta_{\mathrm{fus}} > 0$) or negative (a poorly designed averager that
washes out distinctions will have $\Delta_{\mathrm{fus}} < 0$).

\begin{lemma}[Judge ceiling]
\label{lem:judge-ceiling}
Let $J : (q, (a_{h,q})_h) \to \heads$ be any selector returning one
element of~$\heads$. Then $S(J) \le S(\Omega^*)$, with equality iff
$J(q) \in \arg\max_{h \in \heads} \score(g(q), a_{h,q})$ almost surely
(i.e.\ $J$ picks \emph{some} maximizer at almost every $q$; ties between
heads may be broken arbitrarily). In particular, if there exists an
aggregator~$\Phi$ with $S(\Phi) > S(\Omega^*)$, then no selector
among~$\heads$ matches~$S(\Phi)$.
\end{lemma}

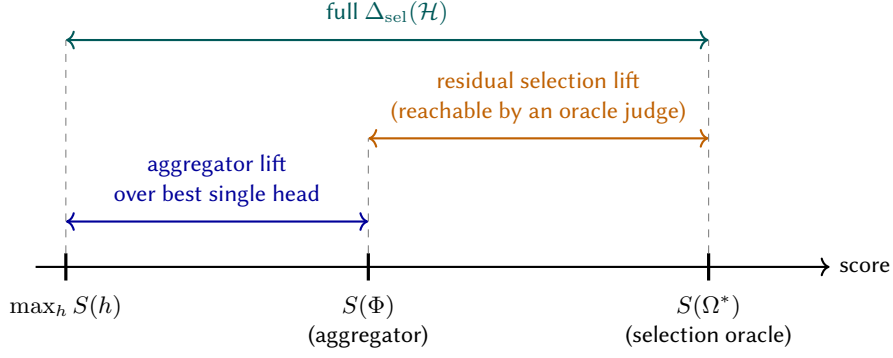
\begin{figure*}[t]
\centering
\begin{tikzpicture}[
  every node/.style={font=\small},
  bar/.style={draw, very thick}
]
  \def\xa{0}\def\xb{4.0}\def\xc{8.5}\def\xe{9.5}
  \def\yblue{0.6}     
  \def\yorange{1.7}   
  \def\yteal{3.0}     

  \draw[->, thick] (-0.4,0) -- (\xe + 0.6,0)
    node[right, font=\footnotesize] {score};
  \foreach \x in {\xa, \xb, \xc} {
    \draw[bar] (\x,-0.18) -- (\x,0.18);
  }
  \node[below=2pt of {(\xa,-0.18)}, anchor=north, font=\footnotesize, align=center]
    {$\max_h S(h)$};
  \node[below=2pt of {(\xb,-0.18)}, anchor=north, font=\footnotesize, align=center]
    {$S(\Phi)$\\(aggregator)};
  \node[below=2pt of {(\xc,-0.18)}, anchor=north, font=\footnotesize, align=center]
    {$S(\Omega^*)$\\(selection oracle)};
  \draw[<->, thick, blue!60!black] (\xa,\yblue) -- (\xb,\yblue)
    node[midway, above=2pt, font=\footnotesize, blue!60!black,
         align=center]
       {aggregator lift\\over best single head};
  \draw[<->, thick, orange!75!black] (\xb,\yorange) -- (\xc,\yorange)
    node[midway, above=2pt, font=\footnotesize, orange!75!black, align=center]
       {residual selection lift\\(reachable by an oracle judge)};
  \draw[<->, thick, teal!70!black] (\xa,\yteal) -- (\xc,\yteal)
    node[midway, above=2pt, font=\footnotesize, teal!70!black]
       {full $\Delta_{\mathrm{sel}}(\heads)$};
  \draw[dashed, gray] (\xa,0.18) -- (\xa,\yteal);
  \draw[dashed, gray] (\xb,0.18) -- (\xb,\yorange);
  \draw[dashed, gray] (\xc,0.18) -- (\xc,\yteal);
\end{tikzpicture}
\caption{\justifying Decomposition of ensemble lift over a candidate
set~$\mathcal{H}$. The aggregator $\Phi$ exceeds the strongest single
head by combining candidate outputs; the residual gap to the selection
oracle is the maximum lift any candidate-restricted selector could add.
On fusion-friendly metrics $\Phi$ can move to the right of
$S(\Omega^*)$ (positive $\Delta_{\mathrm{fus}}$), opening a region no
candidate-restricted selector can reach. On selection-dominated
metrics $S(\Phi) \le S(\Omega^*)$ for any aggregator restricted to the
same candidates, so a judge is the correct combiner.}
\label{fig:decomposition}
\end{figure*}

\noindent The proof is one line: pointwise
$\score(g(q), a_{J(q),q}) \le \max_h \score(g(q), a_{h,q})$, and taking
expectations gives $S(J) \le S(\Omega^*)$, with equality only when $J$
attains the maximum almost everywhere. The lemma is trivial; its
consequence is not. An LLM-as-judge that returns
one candidate verbatim is, by construction, a selector, so its expected
score is a learned approximation of $S(\Omega^*)$. On any metric where a
constructible $\Phi^*$ exceeds $\Omega^*$, no amount of judge tuning closes
the gap. The combiner must include an aggregator, or it is structurally
limited. Figure~\ref{fig:decomposition} illustrates the resulting score
geometry: on fusion-friendly metrics an aggregator $\Phi$ can sit to the
right of the selection oracle $S(\Omega^*)$, opening a region no selector
can reach.

\paragraph{Which BioASQ tracks admit fusion lift?}
\label{sec:answering:syn:taxonomy}

The split is determined by the score function. Two patterns generate
positive $\Delta_{\mathrm{fus}}$: candidate-set-intersection scores
(factoid rank-1 against gold synonyms, the synonym-aware factoid
evaluation style of TREC QA \citep{voorhees2003trecqa}; unioning rank-1
sets monotonically raises hit probability) and recall-oriented
item-level matching scores (list~$F_1$; item-level union raises recall
with a precision trade-off).
Two patterns generate selection dominance \emph{for aggregators
restricted to head candidates}: single-candidate-vs-reference scores
($\score(g, a) = \max_r \mathrm{sim}(a, r)$, e.g., multi-reference
ROUGE) and hard-label indicators (yes/no). Under that restriction
$S(\Omega^*)$ caps every aggregator; an aggregator that synthesises a
new candidate outside~$\heads$ (e.g.\ a generative judge that blends
phrases from multiple heads) could in principle exceed the oracle on
these metrics, but is outside the candidate-restricted setting we
analyse here.
The four BioASQ Phase~B tracks therefore split cleanly: \textbf{YN}
(hard-label) and \textbf{summary/ideal} (multi-reference ROUGE) are
selection-dominated; \textbf{factoid} (rank-1 vs gold synonym set) and
the \emph{recall component} of \textbf{list} (synonym-aware item-level
matching) are fusion-friendly. The list-$F_1$ metric inherits the recall
lift only when the aggregator does not also depress precision --- a
head- and data-dependent condition that, on our Task~13B 2025 heads,
holds for recall but not for $F_1$ (\S\ref{sec:exp:test13}).
Table~\ref{tab:taxonomy} summarises the predicted combiner per track
alongside the empirical confirmation.

\begin{table*}[t]
\centering
\small
\setlength{\tabcolsep}{4pt}
\caption{BioASQ Phase~B metric structure and predicted combiner.}
\label{tab:taxonomy}
\begin{tabular}{p{2.2cm}p{4.3cm}p{3.7cm}p{4.3cm}}
\toprule
Track & Metric structure & Predicted combiner & Empirical (this paper) \\
\midrule
Yes/No & hard-label indicator
& selection (or trivially tied)
& identical across all combiners on val and live (\S\ref{sec:exp:val},
\ref{sec:exp:live}) \\
Factoid & rank-1 set vs synonym gold set
& aggregator (synonym union)
& Phase~A+ b2 factoid MRR tied~\#1 (\S\ref{sec:exp:live}) \\
List & synonym-aware $F_1$ with item-level matching; recall is monotone in
synonym union, precision is not
& aggregator with synonym union (for recall); precision/$F_1$ head-dependent
& synonym-union resolver $+0.025$ \textbf{recall} on test~13B
(\S\ref{sec:exp:test13}); $F_1$ favours GPT-5.5 solo on the same heads
(Table~\ref{tab:test13}); b3 ideal~\#1 (\S\ref{sec:exp:live}) \\
Summary / Ideal & multi-reference ROUGE-2/SU4
& selection dominates
& always-pick floor caught judge regression
(\S\ref{sec:answering:syn:floor}) \\
\bottomrule
\end{tabular}
\end{table*}

The taxonomy puts Lemma~\ref{lem:judge-ceiling} to work. On
selection-dominated metrics an LLM judge can in principle reach the
available headroom. On the recall component of fusion-friendly metrics
it cannot, and the combiner must include an aggregator that exploits
the metric.

\paragraph{The always-pick floor diagnostic.}
\label{sec:answering:syn:floor}

\begin{observation}[Always-pick floor]
\label{obs:always-pick}
The constant selector $\alwayspick{h^{\dagger}}$ that returns
head~$h^{\dagger}$ on every question scores $S(h^{\dagger})$. For any
selector~$J$ to be useful it must satisfy
\(
S(J) \,>\, \max_{h \in \heads} S(\alwayspick{h}).
\)
The diagnostic is $O(|\heads|)$ to compute and routinely catches
calibration errors in deployed LLM judges.
\end{observation}

On Task~13B 2025 with a head set of \{Claude Opus~4.6, Gemini~2.5~Pro,
GPT-5.4\} we measured an LLM-as-judge falling below the floor on two
tracks: on factoid by $0.006$~MRR (judge $0.672$ vs.\
$\max_h S(\alwayspick{h}) = 0.678$) and on summary by $0.006$~ROUGE-2
(judge $0.459$ vs.\ $0.465$). Both regressions were caught within seconds
by Observation~\ref{obs:always-pick} and replaced with deterministic
alternatives before final submission. We carry both design choices into
the current system: the synonym-union resolver on factoid and list (the
fusion-friendly tracks), and a constant-head pick on the selection-dominated summary track.

\paragraph{Implementation.}
\label{sec:answering:syn:impl}

Given list-answer sources from Claude solo
(\S\ref{sec:answering:prompt}), the Gemini 3-stage team
(\S\ref{sec:answering:team}), and GPT-5.5 solo, the resolver runs:
\begin{enumerate}[topsep=2pt,itemsep=1pt,label=(\roman*)]
\item For each question, pool all $(\text{item synonyms},
\text{source})$ tuples.
\item Build equivalence classes by overlapping normalised synonym
strings.
\item Order classes by source-vote count (3-way agreement first, then
2-way, then 1-source).
\item Per class, take the union of all synonyms and pick a canonical
rank-1 form (Claude's first form preferred for canonicality).
\item Cap at 12 items per question.
\end{enumerate}
The output for each question is not, in general, any individual head's
answer: it is a new candidate constructed by step~(iv)'s union --- the
fusion content the lemma identifies. On a 20-question b4 sample the
resolver emits 183 total items (vs.\ Claude 164~/ Gemini 139~/ GPT-5.5
178), with mean 6.21 synonyms per item (vs.\ $\approx 3.5$ for any
individual source) at $+9\%$ item-count growth. The \emph{within-item}
synonym union is precision-preserving (every extra synonym is a fresh
chance of matching a gold form); the \emph{cross-item} growth, however,
admits items proposed by only one head that are sometimes wrong, so the
$F_1$ effect is head-dependent. The empirical confirmation that this beats
every \emph{individual head} on recall is in
\S\ref{sec:exp:test13}: the resolver's list recall on Task~13B 2025 is
$0.653$ vs.\ the best single head's $0.644$. (We do not compute the
per-question selection oracle on list recall, so we cannot claim the
resolver exceeds $S(\Omega^*)$; the empirical claim is the weaker one
that the resolver dominates every constant selector observed.) On
$F_1$ the resolver does not even dominate every individual head ---
GPT-5.5 solo wins list~$F_1$ on the same heads
(Table~\ref{tab:test13}), consistent with the head-dependent precision
behaviour just described.

\section{Experiments}
\label{sec:experiments}

We report three independent evaluations: a controlled retrieval-policy
ablation on Task~12B 2024 validation (\S\ref{sec:exp:val}), an end-to-end
Phase~B comparison on the BioASQ-released gold-input pool of Task~13B
2025 (\S\ref{sec:exp:test13}), and the live preliminary leaderboard for
Task~14B 2026 (\S\ref{sec:exp:live}).

\subsection{Validation: Task 12B 2024 ($n=340$)}
\label{sec:exp:val}

The retrieval-policy ablation is in Table~\ref{tab:ablation}
(\S\ref{sec:reretrieval:full}). The four Phase~B combiners on the same
validation pool (all on Pipeline~B retrieval; full table in
Appendix~\ref{app:val-phaseB}) show three patterns. Claude solo
dominates validation on every metric except list~$F_1$ / list precision
(where Gemini's hybrid team wins by $+0.016$ over Claude solo); the
model rank-order is \emph{input-dependent}: with our retrieved snippets
Claude beats GPT-5.5 on factoid and yes/no by $0.05$--$0.07$, but the
order reverses on gold-quality inputs (\S\ref{sec:exp:test13}). The
synonym-union resolver is the highest-recall list configuration on
validation ($0.5665$, $+0.047$ over Claude solo) and also wins list
recall on test~13B, confirming the framework prediction that the
recall component of fusion-friendly list metrics admits aggregator
lift no selector can reach. The corresponding list-$F_1$ result is
\emph{not} an unconditional aggregator win: GPT-5.5 solo wins list~$F_1$
on Task~13B because the resolver's wider item set costs precision (see
Table~\ref{tab:test13} below).

\subsection{Held-out test: Task 13B 2025 ($n=340$)}
\label{sec:exp:test13}

We evaluate Phase~B answering on the BioASQ-released gold-input pool
(testset 1+2+3+4 merged, $n=340$). This is the same input format the
official Phase~B evaluation uses, so the numbers are directly comparable
to last-year leaderboard entries. We compare five Phase~B configurations:
\textsf{claude\_solo}, \textsf{gpt55\_solo}, \textsf{gemini\_hybrid},
\textsf{claude\_qa\_team}, and the synonym-union resolver
\textsf{resolver} (Table~\ref{tab:test13}).

\begin{table*}[t]
\centering
\small
\caption{Phase~B combiners on Task~13B 2025 BioASQ-released gold-input
pool ($n=340$). \textbf{Bold} = winner per row.}
\label{tab:test13}
\begin{tabular}{llccccc}
\toprule
Type & Metric & claude\_solo & gpt55\_solo & gemini\_hybrid & qa\_team & resolver \\
\midrule
yesno ($n=82$)
& accuracy & 0.9259 & 0.9146 & 0.9390 & \textbf{0.9506} & 0.9259 \\
factoid ($n=95$)
& MRR & 0.7009 & \textbf{0.7044} & 0.6895 & 0.6723 & 0.7009 \\
& strict & 0.6421 & \textbf{0.6737} & 0.6632 & 0.6316 & 0.6421 \\
& lenient & \textbf{0.7684} & 0.7474 & 0.7263 & 0.7263 & \textbf{0.7684} \\
list ($n=83$)
& $F_1$ & 0.6053 & \textbf{0.6518} & 0.6094 & 0.5854 & 0.6194 \\
& precision & 0.6395 & \textbf{0.7070} & 0.6916 & 0.6074 & 0.6528 \\
& recall & 0.6279 & 0.6440 & 0.5855 & 0.6264 & \textbf{0.6528} \\
all ($n=339$)
& ROUGE-2 F & \textbf{0.4623} & 0.4041 & 0.4025 & 0.4581 & \textbf{0.4623} \\
& ROUGE-SU4 F & \textbf{0.4409} & 0.3855 & 0.3784 & 0.4366 & \textbf{0.4409} \\
\bottomrule
\end{tabular}
\end{table*}

\paragraph{Three headline findings.}
\textbf{(1)~No single configuration wins all metrics.} Best per metric is
split four ways: \textsf{qa\_team} (YN); \textsf{gpt55\_solo} (factoid
MRR/strict, list~$F_1$/precision);
\textsf{claude\_solo}/\textsf{resolver} tied (factoid lenient);
\textsf{resolver} (list recall). \textbf{(2)~The team architecture wins YN but loses
lists on gold inputs}, opposite of validation. \textsf{qa\_team} wins
YN by $+0.025$ over Claude solo but regresses list~$F_1$ by $0.020$, and
\textsf{gemini\_hybrid} loses list~$F_1$ to GPT-5.5 solo by
$0.042$; the team's structured evidence brief is helpful compression on
noisy input and lossy on clean input. \textbf{(3)~GPT-5.5 trades
ROUGE for exact-answer $F_1$:} it dominates list precision and factoid
strict via short canonical entity names, while Claude wins ROUGE-2 by
$0.058$ via its verbatim-phrasing prompt. The aggregate ROUGE-2 gap
masks a per-type split: GPT-5.5 wins summary-only ROUGE-2 but loses
yes/no ROUGE-2, because BioASQ's ROUGE scorer reads the full
\texttt{ideal\_answer} field for every question type, and GPT-5.5's
concise yes/no rationales drag its aggregate down.

\paragraph{Significance and leaderboard context.}
Paired bootstrap 95\% CIs ($n_\text{iter}=2000$) on each variant minus
\textsf{claude\_solo} baseline yield nine significant deltas
(Table~\ref{tab:test13-sig}). The two headline rows:

\begin{table*}[t]
\centering
\small
\caption{Significant $\Delta$ vs.\ \textsf{claude\_solo} on Task~13B 2025
(paired bootstrap 95\% CIs, $n_\text{iter}=2000$).}
\label{tab:test13-sig}
\begin{tabular}{llrll}
\toprule
Variant & Metric & Mean $\Delta$ & 95\% CI & Direction \\
\midrule
gpt55\_solo        & list $F_1$        & $+0.0465$ & $[+0.0220, +0.0705]$ & sig.\ better \\
gpt55\_solo        & list precision    & $+0.0675$ & $[+0.0328, +0.1045]$ & sig.\ better \\
resolver          & list recall       & $+0.0249$ & $[+0.0090, +0.0450]$ & sig.\ better \\
gpt55\_solo        & ROUGE-2 F         & $-0.0584$ & $[-0.0749, -0.0419]$ & sig.\ worse \\
gpt55\_solo        & ROUGE-SU4 F       & $-0.0557$ & $[-0.0722, -0.0389]$ & sig.\ worse \\
gemini\_hybrid    & list recall       & $-0.0494$ & $[-0.0929, -0.0134]$ & sig.\ worse \\
gemini\_hybrid    & ROUGE-2 F         & $-0.0600$ & $[-0.0764, -0.0439]$ & sig.\ worse \\
gemini\_hybrid    & ROUGE-SU4 F       & $-0.0626$ & $[-0.0793, -0.0463]$ & sig.\ worse \\
claude\_qa\_team  & list precision    & $-0.0321$ & $[-0.0683, -0.0006]$ & sig.\ worse \\
\bottomrule
\end{tabular}
\end{table*}

the synonym-union resolver wins list recall by $+0.025$, CI
$[+0.009, +0.045]$ (confirming the framework prediction), and GPT-5.5
trades ROUGE-2 ($-0.058$, CI $[-0.075, -0.042]$) for list~$F_1$
($+0.046$, CI $[+0.022, +0.070]$) in a statistically reliable way. Against the published Task~13B 2025 per-batch peak envelope
\citep{nentidis2025bioasq13} --- the per-metric peak averaged across the
four batches, which strictly upper-bounds any individual 2025
participant --- our system sits well above the envelope on the three
factoid metrics, is roughly even on list~$F_1$, and slightly below on
yes/no, where every batch admitted a $1.000$ peak
(Table~\ref{tab:13b-leaderboard}). The system uses 2026-era foundation
models whereas 2025 participants worked with 2024-era models, so the
comparison conflates methodology with base-model strength.

\begin{table*}[t]
\centering
\small
\caption{Our system on Task~13B 2025 vs.\ the published 2025 leaderboard
peak envelope. ``Per-batch range'' is the spread of the per-metric peak
across the four batches; ``Mean of peaks'' averages those four.
``Ours'' is the same metric on the concatenated 340-question test set
with the official synonym-aware scorer. $\Delta$~$=$~Ours~$-$~Mean of
peaks. The envelope strictly upper-bounds any individual 2025
participant.}
\label{tab:13b-leaderboard}
\begin{tabular}{lcccr}
\toprule
Metric & Per-batch range & Mean of peaks & Ours & $\Delta$ \\
\midrule
Yes/No accuracy          & 0.941--1.000 & 0.985 & 0.939 & $-0.046$ \\
Factoid strict accuracy  & 0.450--0.741 & 0.591 & 0.705 & $+0.114$ \\
Factoid lenient accuracy & 0.615--0.815 & 0.702 & 0.789 & $+0.087$ \\
Factoid MRR              & 0.513--0.741 & 0.622 & 0.739 & $+0.117$ \\
List $F_1$               & 0.546--0.591 & 0.567 & 0.578 & $+0.011$ \\
\bottomrule
\end{tabular}
\end{table*}

Even after discounting the 2024$\to$2026 foundation-model uplift, the
factoid margins ($+0.087$ to $+0.117$) remain large; the list-$F_1$
near-tie matches the precision-bounded resolver behaviour reported
above.

On the summary track, the official Perl ROUGE-1.5.5 scorer is not
directly comparable to our max-over-references internal scorer, so we
report a within-pool delta instead: the always-pick rule on a
focused-prompt summary head gives a $+0.080$ ROUGE-2 lift over the
general-prompt single head ($0.385 \to 0.465$) --- the same
head-replacement design \S\ref{sec:answering:syn:floor} flagged on
Task~13B with the prior head set.

\paragraph{Validation\,$\to$\,test-gold gap (claude\_solo only).}
The cleanest evidence for retrieval-vs-answering attribution is the
per-metric val\,$\to$\,test-gold gap on a single configuration that
holds prompt and model fixed (full table in
Appendix~\ref{app:val-to-test}). Yes/no is at the answer ceiling
($\Delta = +0.004$, within run-to-run noise of $\approx 0.02$--$0.03$).
For list~$F_1$ the $+0.132$ absolute gap is consistent with
substantial retrieval-side headroom: prompt and model are held fixed,
and the snippet source differs (our retrieval on val vs.\
BioASQ-released gold inputs on test~13B). The question sets also
differ between Task~12B 2024 and Task~13B 2025, so the gap is not a
clean retrieval-only attribution; it is an upper bound on the lift a
better retrieval pipeline could plausibly recover, not a guarantee.
Factoid MRR shows a smaller retrieval-side gap ($+0.090$) and a larger
remaining prompt-side margin. We treat the retrieval gap as the
highest-priority follow-up because it is large in absolute terms and
because the controlled retrieval ablation in \S\ref{sec:exp:val}
demonstrates the same direction at smaller magnitude.

\subsection{Live test: Task 14B 2026 ($n=240$)}
\label{sec:exp:live}

\paragraph{Submission strategy.}
We submitted under team identifier \textsf{MedQA} on all four batches of
Task~14B, with up to five candidates per phase per batch. The five slots
cover three axes of variation (retrieval source, answer model, and
list method) rather than near-replicas. Slot conventions evolved
across batches: b1 swept Phase~B prompt versions plus a GPT-5.4 hedge
and a combination submission; b2 swept post-processing strategies
(type-selective hybrid, ROUGE-maximising, recall-maximising, Claude or
GPT-5.4 on B+ enriched snippets); b3 and b4 swept retrieval source
$\times$ list method. The slot index \textsf{MedQA-$N$} does not refer
to the same system across batches; we therefore annotate every
leaderboard cell with the internal variant name. The preliminary
leaderboard at the time of writing reports only exact and ideal answers;
Phase~A+ documents-MAP and snippets-MAP tables are pending. Ideal-answer
scores below are ROUGE-SU4~$F_1$ only; official manual ideal-answer
ranks are released post-workshop.

\paragraph{Headline competitive standing.}
\label{sec:exp:live:headline}
The headline result is the combined-exact aggregate: our
team places \emph{first} on the combined-exact aggregate on three of
the eight (phase~$\times$~batch) leaderboards, with second-place
placements on three further leaderboards. At the
question-type level we additionally win or tie-win four individual
cells (factoid on A+~b1, A+~b2, B~b2; ideal on B~b3). The weakest
batch is Phase~B~b4, where no MedQA submission reached top-10 on any
track. Table~\ref{tab:live-headline} summarises the first-place and top-three
placements; the b4 rows are included to reflect the weakest
batch honestly.

\begin{table}[t]
\centering
\small
\setlength{\tabcolsep}{4pt}
\caption{MedQA placements on the Task~14B 2026 preliminary leaderboard.
\textbf{Bold} ranks are first-place auto-metric finishes; ``$>$10''
denotes outside the top~10 (included for b4 to reflect our weakest
batch honestly). When two rows share (Phase, Batch, Metric) they are
distinct submission slots. CSA-IISR (Phase~B lists) and dmiip2024
(Phase~A+ ideal) are the principal competitor families, each winning
the cells we did not.}
\label{tab:live-headline}
\begin{tabular}{@{}cclrr@{}}
\toprule
Ph & B & Metric & Rank & Score \\
\midrule
A+ & 1 & factoid (MRR)        & \textbf{1}       & 0.478 \\
A+ & 1 & factoid (MRR)        & 2                & 0.435 \\
A+ & 2 & factoid (MRR)        & \textbf{1} (tie) & 0.400 \\
A+ & 2 & factoid (MRR)        & 2                & 0.400 \\
A+ & 2 & combined exact       & \textbf{1}       & 0.567 \\
A+ & 2 & combined exact       & 2                & 0.564 \\
A+ & 3 & combined exact       & \textbf{1}       & 0.576 \\
A+ & 3 & combined exact       & 2                & 0.574 \\
A+ & 4 & combined exact       & 3                & 0.533 \\
A+ & 4 & combined exact       & 4                & 0.529 \\
A+ & 4 & list $F_1$           & 7                & 0.415 \\
A+ & 4 & list $F_1$           & 8                & 0.403 \\
B  & 1 & ideal (R-SU4 $F_1$)  & 2                & 0.233 \\
B  & 1 & ideal (R-SU4 $F_1$)  & 3                & 0.233 \\
B  & 2 & factoid (MRR)        & \textbf{1} (tie) & 0.400 \\
B  & 2 & combined exact       & \textbf{1}       & 0.596 \\
B  & 3 & ideal (R-SU4 $F_1$)  & \textbf{1}       & 0.255 \\
B  & 3 & ideal (R-SU4 $F_1$)  & 3                & 0.247 \\
B  & 3 & combined exact       & 4                & 0.672 \\
B  & 4 & list $F_1$           & $>$10            & 0.486 \\
B  & 4 & combined exact       & $>$10            & 0.557 \\
\bottomrule
\end{tabular}
\end{table}

\paragraph{Out-of-sample variant ablation.}
Because each batch's five slots cross our diversification axes, the live
leaderboard doubles as an out-of-sample ablation, from which several
findings stand out.

\paragraph{Adding the hybrid first stage is fusion-friendly on ideal but
selection-dominated on exact.} On Phase~B~b3, \textsf{crossmodel}
(agent~$\cup$~hybrid first-stage; Gemini list team) wins ideal at
R-SU4~$F_1 = 0.255$ (\#1) while losing combined exact by $0.002$ to the
simpler agent-only $c_1$ ($0.576$ vs $0.574$). The two variants share the
same Claude summary head, so the ROUGE delta originates upstream in the
broader candidate pool the ideal-answer head summarises. For combined
exact, the BGE rerank-and-top-$N$ cap occasionally promotes a
near-duplicate snippet over a higher-precision candidate. The reading
matches \S\ref{sec:answering:syn:taxonomy}'s prediction: ensembling
widens the entity inventory the ideal track rewards, but past a per-type
optimum it dilutes the gestalt rank-1 signal exact answers depend on.

\paragraph{Claude beats GPT under matched prompts on factoid, twice.}
Two clean model-only ablations on Phase~B with identical single-agent prompts.
In~b1, MedQA-4 (\textsf{gpt54}) lagged the four Claude variants on
factoid MRR ($0.449$ vs $0.493$). In~b4,
MedQA-2 (\textsf{gpt55\_solo}) was the weakest of our five submissions on
combined exact. Both swaps left everything else fixed: the gap is
attributable to the model. Combined with the b2 result that the
type-selective \textsf{hybrid} variant (MedQA-1) wins combined exact and
ties \#1 on factoid MRR, the live data supports a routing-based system
over a single-model fallback.

\paragraph{b4 list~$F_1$ partially diagnoses our 3-stage list team.}
b4 is our weakest batch: no MedQA submission reached top-10 combined
exact. YN ($0.867$) and factoid ($0.318$) were typical; list~$F_1$ was
the localised gap, where our best (\textsf{crossmodel}, $0.486$) trailed
CSA-IISR ($0.627$--$0.654$). Direct measurement on the 20 b4 list
submissions explains part: mean synonyms per item are $3.99$ for Claude
solo, $3.48$ for the Claude 3-stage team (\textsf{hybrid}), and
$3.49$ for the Gemini team (\textsf{crossmodel}), a 12--13\%
synonym compression at the Validator's entity-normalisation pass. The
synonym-union resolver (\textsf{super\_crossmodel}) recovers synonym
depth ($6.21$/item) but underperforms \textsf{crossmodel} on b4 $F_1$
($0.442$ vs $0.486$) because it admits items proposed by only one head
at higher rates ($9.15$ items/q vs $6.95$), some of which are wrong. CSA-IISR's
mechanism for outperforming both combiners is not visible from our data.

\paragraph{Phase~A+ b4: indirect evidence for the hybrid first stage.}
Phase~A+~b4 placements are also below MedQA's typical range, with the
best combined-exact at \#3 and factoid and list outside the top-5.
Three of our five intended b4 Phase~A+ submissions, all of which used
the hybrid first stage in addition to the agent pipeline, suffered
upload-pipeline failures shortly before the BioASQ deadline, and the
fallback submissions that were posted on time used the agent-retrieval
pool alone (without the hybrid first-stage ensemble). This was
unintentional, but it produced a natural ablation: in b2 and b3, where
the hybrid-enabled variants were submitted, our system placed
\emph{first} on combined-exact; in b4, where they were not, our best
placement on combined-exact dropped to \#3 and the system fell out of
the top-10 on factoid and list. The delta is consistent with the
framing in
\S\ref{sec:retrieval:hybrid}--\ref{sec:retrieval:ensemble}: the hybrid
first stage is a substantive contributor to Phase~A+ standing rather
than a redundant addition to the agent pipeline.

\section{Discussion}
\label{sec:discussion}

\subsection{Cost-pragmatic gains, retrieval-robust YN, and the
metric-structure principle}
\label{sec:disc:findings}

\paragraph{Cost-pragmatic wins on lists; YN is retrieval-robust.}
The mechanism behind Pipeline~B's list-$F_1$ gain is the
union-rerank-cap structure: BGE rerank-then-top-$N$ caps the pool, so
rescue re-retrieval can only displace lower-quality snippets, never
inflate the pool. YN does not benefit because a single best snippet is
enough; the four-category framework converges deterministically given
$\ge \!\sim\!10$ relevant sentences. Across every retrieval and answering
variant tested, paired YN answers were identical, and the remaining
$8/102$ val errors trace to genuinely ambiguous gold labels or fuzzy
Category-B/C boundaries, consistent with the
\S\ref{sec:answering:syn:taxonomy} prediction that YN is a
selection-dominated metric.

\paragraph{Synonym union is BioASQ's free lunch on list \emph{recall}.}
Synonym-aware list scoring is monotone in the synonym set, so unioning
\emph{within-item} synonyms across independent heads is a
precision-preserving recall gain. The three-head resolver raises mean
synonyms per item from $\approx 3.5$ to $6.21$ at $+9\%$ items, because
the three heads express the same entity in different canonical forms
(``KRAS G12C''~/ ``G12C''~/ ``KRAS p.G12C'') --- exactly the forms gold
uses. The $+9\%$ \emph{cross-item} growth, however, is not
precision-preserving: it admits items proposed by only one head that
depress precision, so the resolver wins list recall but loses list~$F_1$ to
GPT-5.5 solo on Task~13B 2025 (\S\ref{sec:exp:test13},
Table~\ref{tab:test13}). The wider conclusion follows from
Lemma~\ref{lem:judge-ceiling}: the choice between an LLM judge and an
aggregator should be driven by the metric \emph{and} by the empirical
precision behaviour of the aggregator on the heads at hand --- the
$F_1$ benefit is conditional, not automatic.

\subsection{The always-pick floor as a deployment diagnostic}
\label{sec:disc:floor}

Lemma~\ref{lem:judge-ceiling} has an operational corollary. A
multi-candidate LLM judge $J$ whose deployed score falls below
$\max_h S(\alwayspick{h})$ is destroying value relative to a constant
rule that requires no inference. The diagnostic is $O(|\heads|)$ to
compute and frequently violated, because a competently-prompted judge
over-trusts a strong head and loses to the constant rule on the
residual. We carry two design choices from this diagnostic: list
answers go through the synonym-union resolver (judge fell below the
floor on factoid by $0.006$~MRR in our prior measurement on Task~13B
2025), and summary answers go through a single focused-prompt Claude
head (judge fell below the floor by $0.006$~ROUGE-2 on the same
measurement). We recommend this diagnostic accompany every deployed
multi-candidate judge.

\subsection{Limitations and future work}
\label{sec:disc:limits}

\paragraph{Limitations.}
Per-type validation $n$ is small ($80$--$102$), so non-inferiority CIs
on factoid cross zero at $n=85$; replicators should plan for
$n\!\ge\!150$. Phase~A retrieval on test~13B was not run, so the
$+0.132$ list-$F_1$ val\,$\to$\,test-gold gap is a lower bound on
retrieval-recoverable headroom. ROUGE numbers use an internal
max-over-references scorer not directly comparable to the official Perl
ROUGE-1.5.5 \citep{lin2004rouge} in absolute terms; within-system
deltas remain valid. Run-to-run variance on agent queries is unmeasured
($\approx 0.02$--$0.03$ factoid~MRR across replicates); reported CIs are
paired-bootstrap across questions, holding the agent trajectory fixed.
Phase~A+ documents/snippets MAP and final manual ideal-answer scores
for Task~14B are pending release, so the live-leaderboard standing
reported here is the answering-side picture only; the b3 ideal~\#1
placement is preliminary on the auto-ROUGE axis. Between-team
comparison should use each team's best-of-five slot (BioASQ allows five
submissions per team, and Table~\ref{tab:live-headline} rows are
MedQA's best of five).

\paragraph{Future work.}
Three directions follow. \textbf{(i)}~A constructive LLM judge that
performs synonym union when heads agree on an entity but disagree on
surface form, and item-level union when heads partially agree on a
list, extends LLM-as-judge to fusion-friendly metrics;
Lemma~\ref{lem:judge-ceiling} bounds single-pick judges but not
constructors. \textbf{(ii)}~Retrieval is the dominant gain budget on
list~$F_1$ ($+0.132$ val\,$\to$\,test-gold gap, plausibly
retrieval-recoverable); denser indexes, better
citation-graph expansion, and learned re-ranking conditioned on
per-question entity inventory are higher-leverage than further
answering-side tuning. \textbf{(iii)}~The b4 list-$F_1$ gap to CSA-IISR
and ku\_dmis ($0.486$ vs $0.627$--$0.654$ on the same 20-question
split) is the single largest competitor delta we exhibit; ablating the
3-stage team's Validator normalisation pass is the most informative
follow-up.

\section{Conclusion}
\label{sec:conclusion}

A BGE quality gate plus selective re-retrieval beats indiscriminate
re-retrieval at lower cost on list~$F_1$ and list precision, with
no significant difference on YN, factoid, or list recall
(Table~\ref{tab:ablation}); the val\,$\to$\,test-gold gap on Task~13B
2025 indicates substantial list-$F_1$ headroom ($+0.132$ absolute)
consistent with retrieval-side improvement, though question sets
differ between val and test~13B so the attribution is not exclusive. A hybrid first stage
(BGE~+~BM25~+~RRF, no cross-encoder reranker;
Table~\ref{tab:rerankers}) ensembled with an agent-driven pipeline
produces nearly disjoint candidates ($\approx 4.5\%$ overlap) and feeds
a richer Phase~A+ pool. For Phase~B, metric structure determines the
combiner: selection-dominated metrics (YN, multi-reference ROUGE) admit
LLM judging, the recall component of fusion-friendly metrics (factoid
rank-1, list \emph{recall} with synonym matching) does not, and our
deterministic synonym-union resolver exceeds every \emph{individual
head} on recall (Lemma~\ref{lem:judge-ceiling} bounds every
candidate-restricted selector; resolver list recall $0.6528$ vs.\ best
single head $0.6440$; $+0.025$ vs.\ the Claude baseline, CI~$[+0.009,
+0.045]$; Phase~B b3 ideal~\#1). On the
precision-bounded list-$F_1$ metric the resolver does not dominate
every selector --- GPT-5.5 solo retains the $F_1$ lead on Task~13B
(Table~\ref{tab:test13}) --- so the framework's prescription for $F_1$
is conditional on the aggregator's precision behaviour, not
unconditional. On the Task~14B 2026
preliminary leaderboard our system places first on the combined-exact
aggregate on three of the eight (phase~$\times$~batch) leaderboards and wins four individual question-type cells (three
factoid, one ideal). The methodological takeaway: LLM judges are
powerful approximators of the selection oracle but are not, by
themselves, an ensembling primitive: the choice between judging and
aggregation should be driven by the metric, not by the toolkit.

\begin{acknowledgments}
We gratefully acknowledge the U.S.\ National Library of Medicine for
PubMed and the NIH Office of Portfolio Analysis for the iCite citation
API
\citep{hutchins2019icite}; EMBL-EBI for Europe~PMC
\citep{levchenko2018europepmc}; the BAAI team for the BGE encoder and
reranker family \citep{chen2024bgem3}; and the BioASQ organisers and
the CLEF community.
\end{acknowledgments}

\section*{Declaration on Generative AI}
During the preparation of this work, the author(s) used Claude
Opus~4.6, Gemini~2.5~Pro, GPT-5.4, and GPT-5.5 as components
of the BioASQ system being described; the language-model outputs
they produce are the experimental subject of this paper. The author(s)
additionally used Claude Opus~4.6 for: drafting assistance, grammar
and spelling check, and code generation in support of the experimental
pipeline. After using these tools, the author(s) reviewed and edited
the content as needed and take full responsibility for the
publication's content.


\clearpage
\appendix

\section{Retrieval-pipeline benchmarks and design studies}
\label{app:retrieval-studies}

This appendix carries the retrieval benchmark numbers, the
hybrid-fusion ablation, and the encoder / cross-encoder reranker
studies that motivate the design in \S\ref{sec:retrieval:hybrid}. All
results are on the BioASQ-13b retrieval slice (3{,}986 questions,
years~5--13) with the rest of the pipeline held fixed.

\paragraph{Headline numbers.}
The hybrid first stage reaches mean R@200 = 99.3\%,
R@50 = 95.0\%, R@10 = 68.7\%, P@10 = 56.2\%, $F_1$@10 = 61.8\%, with
query latency under 100~ms. This evaluation slice is larger and spans more years than the
Task~12B--14B sets used elsewhere in the paper; we report it to characterise the retriever's recall ceiling at
scale rather than as an evaluation of the full system.

\paragraph{Hybrid is what unlocks recall.}
BM25-only and semantic-only retrieval plateau at 80.6\% and
$\approx 83\%$ R@200 respectively, regardless of $K$. The RRF-fused
hybrid reaches 99.3\%. The two retrievers are complementary: roughly
two-thirds of the queries hinge on exact biomedical-term matching that
BM25 handles well, while the remaining third require paraphrase
semantics that the dense encoder supplies. This complementarity is the
single largest design-driven recall gain we identify.

\begin{table*}[t]
\centering
\small
\caption{Embedding-model comparison on the BioASQ-13b retrieval slice.
Higher R@10 is better.}
\label{tab:embeddings}
\begin{tabular}{lcl}
\toprule
Model & R@10\,$\uparrow$ & Note \\
\midrule
\textbf{BAAI/bge-base-en-v1.5}            & \textbf{0.687} & Production choice \\
ncbi/MedCPT (asymmetric Q+A)              & 0.675 & Best biomedical-specific \\
BAAI/bge-large-en-v1.5                    & 0.669 & 1024d --- worse than base \\
abhinand/MedEmbed-base-v0.1               & 0.666 & LLaMA-trained \\
nomic-ai/nomic-embed-text-v1.5            & 0.662 & Best with prefix instructions \\
intfloat/e5-base-v2                       & 0.661 & Fastest 768d model \\
NeuML/pubmedbert-base-embeddings          & 0.623 & Full transformer \\
NeuML/pubmedbert-base-embeddings-8M       & 0.617 & Model2Vec (ultra-fast) \\
\bottomrule
\end{tabular}
\end{table*}

\section{Pilot re-retrieval ablation ($n=100$)}
\label{app:pilot}

The full validation ablation in \S\ref{sec:reretrieval:full} was preceded
by a pilot on 100 validation questions (yes/no~$=30$, factoid~$=25$,
list~$=24$, summary~$=21$). The pilot returned an inconclusive
non-inferiority verdict at this $n$ but two clean directional findings:
B-min beat~A on factoid MRR by $+0.048$ (paired Wilcoxon
$p<\epsilon$), and the rescue step appeared to \emph{hurt} factoid at this
$n$ ($\text{B} < \text{B-min}$ by $0.041$~MRR). Yes/no was already tied at
$0.900$/$0.933$/$0.933$ with 0/30 paired disagreements. The full
ablation at $n=340$ confirmed the direction of the factoid finding with
smaller-but-consistent magnitude (B-min~$>$~A by $+0.022$~MRR) and
reversed the rescue-hurts-factoid direction once the broader question
sample is observed. A pilot at $n\!\approx\!100$ is therefore sufficient
to identify direction but not significance for factoid; replicators
should plan for $n\!\ge\!150$ per condition.

\section{Phase~B combiners on Task~12B 2024 validation}
\label{app:val-phaseB}

The Phase~B combiner comparison on the Task~12B 2024 validation pool
($n=340$, Pipeline~B retrieval; cf.\ \S\ref{sec:exp:val}) appears in
Table~\ref{tab:val-phaseB}. Three patterns are visible: Claude solo
dominates on every metric except list~$F_1$ / list precision (won by
Gemini's hybrid team); the model rank-order is input-dependent, reversing
on gold-quality inputs in Task~13B (\S\ref{sec:exp:test13}); and the
synonym-union resolver is the highest-recall list configuration on
validation. The resolver inherits Claude's non-list answers and is
therefore identical to \textsf{claude\_solo} on yes/no, factoid, and
ROUGE columns; it differs only on list metrics.

\begin{table*}[t]
\centering
\small
\caption{Phase~B combiners on Task~12B 2024 validation ($n=340$); all on
Pipeline~B retrieval (\S\ref{sec:reretrieval}). \textbf{Bold} = winner
per row.}
\label{tab:val-phaseB}
\begin{tabular}{llcccc}
\toprule
Type & Metric & claude\_solo & gpt55\_solo & gemini\_hybrid & resolver \\
\midrule
yesno ($n=102$)
& accuracy & \textbf{0.9216} & 0.8627 & 0.8333 & \textbf{0.9216} \\
factoid ($n=85$)
& MRR & \textbf{0.6106} & 0.5882 & 0.5327 & \textbf{0.6106} \\
& strict & \textbf{0.5765} & 0.5294 & 0.4881 & \textbf{0.5765} \\
& lenient & \textbf{0.6824} & 0.6588 & 0.6071 & \textbf{0.6824} \\
list ($n=80$)
& $F_1$ & 0.4733 & 0.4644 & \textbf{0.4892} & 0.4830 \\
& precision & 0.4737 & 0.4664 & \textbf{0.5088} & 0.4620 \\
& recall & 0.5192 & 0.5124 & 0.5186 & \textbf{0.5665} \\
all ($n=340$)
& ROUGE-2 F & \textbf{0.2543} & 0.2294 & 0.2193 & \textbf{0.2543} \\
& ROUGE-SU4 F & \textbf{0.2447} & 0.2212 & 0.2097 & \textbf{0.2447} \\
\bottomrule
\end{tabular}
\end{table*}

\section{Validation\,$\to$\,test-gold gap for \textsf{claude\_solo}}
\label{app:val-to-test}

Table~\ref{tab:val-to-test} reports the per-metric gap between our
Task~12B 2024 validation results (with Pipeline~B retrieval) and the
held-out Task~13B 2025 results (on BioASQ-released gold-input snippets),
for a single Phase~B configuration \textsf{claude\_solo} held constant on
both sides. The prompt and model do not change between the two
columns, but the question set differs (Task~12B 2024 vs.\ Task~13B
2025), so the $\Delta$ is consistent with retrieval-side headroom
rather than exclusively attributable to it. Yes/no sits at the answer
ceiling ($+0.004$, within run-to-run noise); list~$F_1$ exhibits the
largest gap at $+0.132$ absolute, which is the plausibly
retrieval-recoverable headroom discussed in \S\ref{sec:exp:test13} and
\S\ref{sec:disc:limits}.

\begin{table*}[t]
\centering
\small
\caption{Validation\,$\to$\,test-gold gap for \textsf{claude\_solo}: same
prompt and model on both sides, only the snippet source changes (our
retrieval on val vs.\ BioASQ-released gold inputs on test~13B).}
\label{tab:val-to-test}
\begin{tabular}{lccc}
\toprule
Metric & Val 12B (Pipeline B) & Test 13B (gold inputs) & $\Delta$ \\
\midrule
YN accuracy     & 0.9216 & 0.9259 & $+0.004$ \\
Factoid MRR     & 0.6106 & 0.7009 & \textbf{$+0.090$} \\
Factoid strict  & 0.5765 & 0.6421 & $+0.066$ \\
Factoid lenient & 0.6824 & 0.7684 & $+0.086$ \\
List $F_1$      & 0.4733 & 0.6053 & \textbf{$+0.132$} \\
List recall     & 0.5192 & 0.6279 & $+0.109$ \\
List precision  & 0.4737 & 0.6395 & $+0.166$ \\
\bottomrule
\end{tabular}
\end{table*}

\section{Extended Phase~B findings on Task~13B 2025}
\label{app:test13-extended}

This appendix carries auxiliary findings from Table~\ref{tab:test13}
that did not fit the main \S\ref{sec:exp:test13} narrative.

\paragraph{Team architecture: helpful on noise, lossy on gold.}
The \textsf{claude\_qa\_team} variant wins yes/no on Task~13B by
$+0.025$ over Claude solo but \emph{regresses} list~$F_1$ by $0.020$;
\textsf{gemini\_hybrid} loses list~$F_1$ to GPT-5.5 solo by
$0.042$. On validation, the same team architecture \emph{improved}
Claude list~$F_1$. The simplest reconciliation is that the team's
Analyst stage compresses noisy snippets in a way that strips noise
but, on gold-quality inputs, also strips legitimate entity-form
variation; the Validator stage's value is consistent regardless (it
catches YN category mis-routing in both regimes).

\section{Per-variant analysis of the Task~14B preliminary leaderboard}
\label{app:live-detail}

\S\ref{sec:exp:live} highlights three live-leaderboard findings; this
appendix carries the other variant-level observations.

\paragraph{b2 Phase~B: type-selective hybrid post-processing.}
\textsf{hybrid} (MedQA-1: Claude-resolved yes/no + cross-model factoid
synonym union + Claude+consensus list union + best-of-3 summary picker)
placed \textbf{first on combined exact} in Phase~B b2 (0.596) and tied
\textbf{first on factoid MRR} (0.400). The more aggressive
\textsf{recall} variant (MedQA-3, full 3-model list union and
aggressive synonym blow-up via an \texttt{expand\_variants()}
post-processor) scored $0.583$ on combined exact, slightly weaker than
\textsf{hybrid}. The pattern matches the validation observation that
synonym union beyond a moderate level adds precision risk faster than
it adds recall, so the best-balanced hedge wins.

\paragraph{b3 Phase~B cross-model routing on ideal: mechanism is
upstream.} The \textsf{crossmodel} variant (MedQA-5: Claude solo for YN,
factoid, summary; Gemini 3-stage team for lists) wins Phase~B b3 ideal
at R-SU4~$F_1 = 0.255$, $+0.008$ above $c_1$ (Claude-only solo at
$0.247$, also podium). Both share the same Claude summary head for the
ideal-answer prose, so the delta originates upstream: either in the
Gemini list team contributing a different entity inventory that the
ideal-answer head subsequently summarises, or in the snippet pool
differing because list-question retrieval was independently merged with
the hybrid first stage for \textsf{crossmodel}. We do not isolate these
two contributions on b3 alone; what the data supports is the direction
without committing to a specific mechanism.

\paragraph{dmiip2024 is the principal Phase~A+ ideal competitor.}
The dmiip2024 system family sweeps the ROUGE-SU4 top-3 on Phase~A+ ideal
in three of four batches. Our \textsf{cascade\_v2m} variant (MedQA-2 in
b2 A+) reached ideal R-SU4 $0.152$, well outside the top-10 for that
batch. The gap is consistent across batches and is the largest
single-axis competitor delta we exhibit on the leaderboard, a direct
improvement target for the ideal-answer prompt. Inspection suggests
dmiip2024's ideal answers are systematically longer and more
snippet-verbatim than ours, which favours the ROUGE-SU4 bigram and
skip-bigram overlap.

\section{The Claude solo prompt: design choices}
\label{app:claude-prompt}

The Claude solo prompt anchors all combiner configurations. The four design
choices below were each validated against an alternative on Task~12B
2024 before being adopted.

\paragraph{Yes/no four-category framework.}
Every yes/no question is first classified into one of four categories:
\textbf{A} (association/mechanism: ``yes if any snippet reports it'');
\textbf{B} (clinical use/benefit: ``yes if any RCT/guideline/cohort
shows it''); \textbf{C} (approval/safety/standard: ``no unless explicit
regulatory or guideline statement''); \textbf{D} (factual claim, match
against snippets). The category letter is the first word of the
\texttt{ideal\_answer}. This eliminates the systematic mis-routing
errors of earlier prompts (false-positive Bs that should have been Cs).

\paragraph{Mandatory five factoid candidates with synonym variants.}
Instead of ``list up to 5 candidates,'' we require exactly five
candidates per question, each with 4--6 synonym variants (gene symbols
+ expansions, abbreviations + full names, spelling variants, definition
forms for ``what is~$X$?'' questions). This outperforms ``up to 5'' on
both strict and lenient factoid accuracy.

\paragraph{Two-stage list extraction with synonym expansion.}
Stage~1 enumerates entities matching the question's abstraction level;
stage~2 generates 4--6 synonyms per item (full + abbreviation +
alternative spellings + brand/generic + American/British). A coverage
re-read after stage~1 adds non-trivial list recall.

\paragraph{Verbatim summary phrasing, 35--50 words / 250--280 chars.}
Match the gold length distribution; copy verbatim phrases from the
strongest snippet to maximise ROUGE-2 bigram overlap. Self-check
instructions on the summary track were tried and hurt because they
caused over-editing of the verbatim copy.

\section{Extended discussion: mechanisms behind the headline findings}
\label{app:disc-extended}

\paragraph{Why list~$F_1$ benefits from rescue but factoid does not.}
List~$F_1$ is harmonic in recall and precision; the rescue step
improves recall by handing more gold items to the answering agent
without proportionally increasing precision-killing noise: the BGE
rerank-and-top-$N$ cap that follows the merge prevents the union from
inflating the per-question pool size, so re-retrieval can only displace
lower-quality snippets, never inflate the pool. Yes/no and factoid do
not get this benefit because they are not set-valued: a single best
snippet is enough, and once retrieval clears the per-question top-1
threshold, additional snippets neither help nor hurt.

\paragraph{Why yes/no is fully retrieval-robust.}
The four-category framework (\S\ref{sec:answering:prompt}) converges
deterministically given a snippet pool with $\ge\!\sim\!10$ relevant
sentences. The remaining 8/102 validation errors are not retrieval
failures but either genuinely ambiguous gold labels (mixed-evidence
Category-C questions where reasonable annotators disagree) or
mis-classified Category~B vs.\ C questions where the prompt's category
boundary is fuzzy. The path forward is prompt design, not retrieval ---
consistent with \S\ref{sec:answering:syn:taxonomy}'s prediction that
yes/no is a selection-dominated metric and any combiner over fixed
candidates is bounded by the per-question selection oracle.

\paragraph{Why the synonym-union ``free lunch'' works (and where it stops).}
The three heads (Claude solo, Gemini 3-stage team, GPT-5.5
single-agent) tend to express the same entity in different canonical
forms (``KRAS G12C'' vs.\ ``G12C'' vs.\ ``KRAS p.G12C''), exactly
the forms gold uses. Because BioASQ list~$F_1$ counts any synonym in
the gold's reference set as a hit, the \emph{within-item} synonym
union is precision-preserving: a synonym that does not match gold
contributes nothing, and one that does adds a recall gain. The same
logic applies to factoid rank-1 against gold synonyms. The
\emph{cross-item} union --- adding items that only one head proposed
--- is not free: such items, when wrong, depress precision, which is why the resolver wins list recall but loses
list~$F_1$ to GPT-5.5 solo on Task~13B (Table~\ref{tab:test13}). The
lemma's bite is that the recall gain itself is fundamentally
unavailable to any LLM-as-judge that returns one candidate verbatim,
regardless of how well that judge is prompted; whether the gain
translates to $F_1$ is a separate, precision-dependent question.

\end{document}